%% file: main.tex
\documentclass{article}

\usepackage{microtype}
\usepackage{graphicx}
\usepackage{subfigure}
\usepackage{booktabs} 
\usepackage{bbding}
\usepackage{amssymb}
\usepackage{amsmath}
\usepackage{pifont}
\usepackage{tabularx}
\usepackage{minted}
\usepackage{caption}
\usepackage{multirow}
\usepackage{algorithm}
\usepackage{wrapfig}

\usepackage[T1]{fontenc}
\usepackage[utf8]{inputenc}

\newcommand{\cmark}{\ding{51}}%
\newcommand{\xmark}{\ding{55}}%

\setminted[python]{ %
    linenos=true,             
    autogobble=true,          
    frame=lines,
    framesep=2mm,
    fontsize=\footnotesize
}


\usepackage{hyperref}

\usepackage[arxiv]{mlsys2023}

\mlsystitlerunning{TreeTensor: Boost AI System on Nested Data with Constrained Tree-Like Tensor}

\begin{document}

\twocolumn[
\mlsystitle{TreeTensor: Boost AI System on Nested Data with Constrained Tree-Like Tensor}



\mlsyssetsymbol{equal}{*}

\begin{mlsysauthorlist}
\mlsysauthor{Shaoang Zhang}{equal,buaa,shlab}
\mlsysauthor{Yazhe Niu}{equal,shlab,cuhk}
\end{mlsysauthorlist}

\mlsysaffiliation{buaa}{School of Computer Science and Engineering, Beihang University, Beijing, China}
\mlsysaffiliation{shlab}{Shanghai Artificial Intelligence Laboratory, Shanghai, China}
\mlsysaffiliation{cuhk}{Multimedia Laboratory, The Chinese University of Hong Kong, Hong Kong, China}

\mlsyscorrespondingauthor{Shaoang Zhang}{hansbug@buaa.edu.cn}
\mlsyscorrespondingauthor{Yazhe Niu}{niuyazhe314@outlook.com}

\mlsyskeywords{Machine Learning, MLSys}

\vskip 0.3in

\input{sections/0.abstract}
]



\printAffiliationsAndNotice{\mlsysEqualContribution} 

\input{sections/1.introduction}
\input{sections/2.related_works}
\input{sections/3.treevalue}
\input{sections/4.examples}

\input{sections/5.limitations}

\input{sections/6.conclusion}

\section*{Acknowledgements}

We thank the reviewers for their valuable feedback and suggestions that helped improve this paper. We also acknowledge the support from our institutions and collaborators who contributed to the development of this work.



\bibliography{example_paper}
\bibliographystyle{mlsys2023}



\end{document}

%% file: sections/0.abstract.tex
\begin{abstract}

%

\textit{Tensor} is the most basic and essential data structure of nowadays artificial intelligence (AI) system.
The natural properties of Tensor, especially the memory-continuity and slice-independence, make it feasible for training system to leverage parallel computing unit like GPU to process data simultaneously in batch, spatial or temporal dimensions.
However, if we look beyond perception tasks, the data in a complicated cognitive AI system usually has hierarchical structures (\textit{i.e.} nested data) with various modalities. They are inconvenient and inefficient to program directly with conventional Tensor with fixed shape. 
To address this issue, we summarize two main computational patterns of nested data, and then propose a general nested data container: \textit{TreeTensor}. 
Through various constraints and magic utilities of TreeTensor, one can apply arbitrary functions and operations to nested data with almost zero cost, including some famous machine learning libraries, such as Scikit-Learn, Numpy and PyTorch. 
Our approach utilizes a constrained tree-structure perspective to systematically model data relationships, and it can also easily be combined with other methods to extend more usages, such as asynchronous execution and variable-length data computation.
Detailed examples and benchmarks show TreeTensor not only provides powerful usability in various problems, especially one of the most complicated AI systems at present: AlphaStar for StarCraftII, but also exhibits excellent runtime efficiency without any overhead. Our project is available at \url{https://github.com/opendilab/DI-treetensor}.

\end{abstract}

%% file: sections/1.introduction.tex
\section{Introduction}
\label{section-introduction}


In recent years, data-driven deep learning methods have made a great progress in many complicated artificial intelligence (AI) applications, such as image-text generation \cite{clip, dalle}, protein structure prediction \cite{alphafold}, human-level chess and video game decision making \cite{alphazero, badia_agent57_2020} and so on. 
With the rapid advances of algorithms, technical toolkits play a more significant role in bridging the gap between academic research and industrial practice. Therefore, various deep learning training and inference frameworks \cite{abadi_tensorflow_2016, paszke_pytorch_2019, chen_mxnet_2015} are continually updated and evolved to reduce cost and increase efficiency for both training and deployment. 
One of the most significant elements in these frameworks is Tensor \cite{abadi_tensorflow_2016}, a regular multi-dimension data structure. The neat data arrangement of Tensor enables rapid and parallel processing on advanced computation devices like GPU \cite{gpu} and TPU \cite{tpu}. AI model, usually neural network, can be implemented through corresponding flexible programming models and interfaces based on Tensor. Moreover, users can simply describe the data flow of Tensor to construct and execute powerful neural network in highly parallel. 

\begin{figure*}[t]
    \centering
    \includegraphics[width=0.9\textwidth]{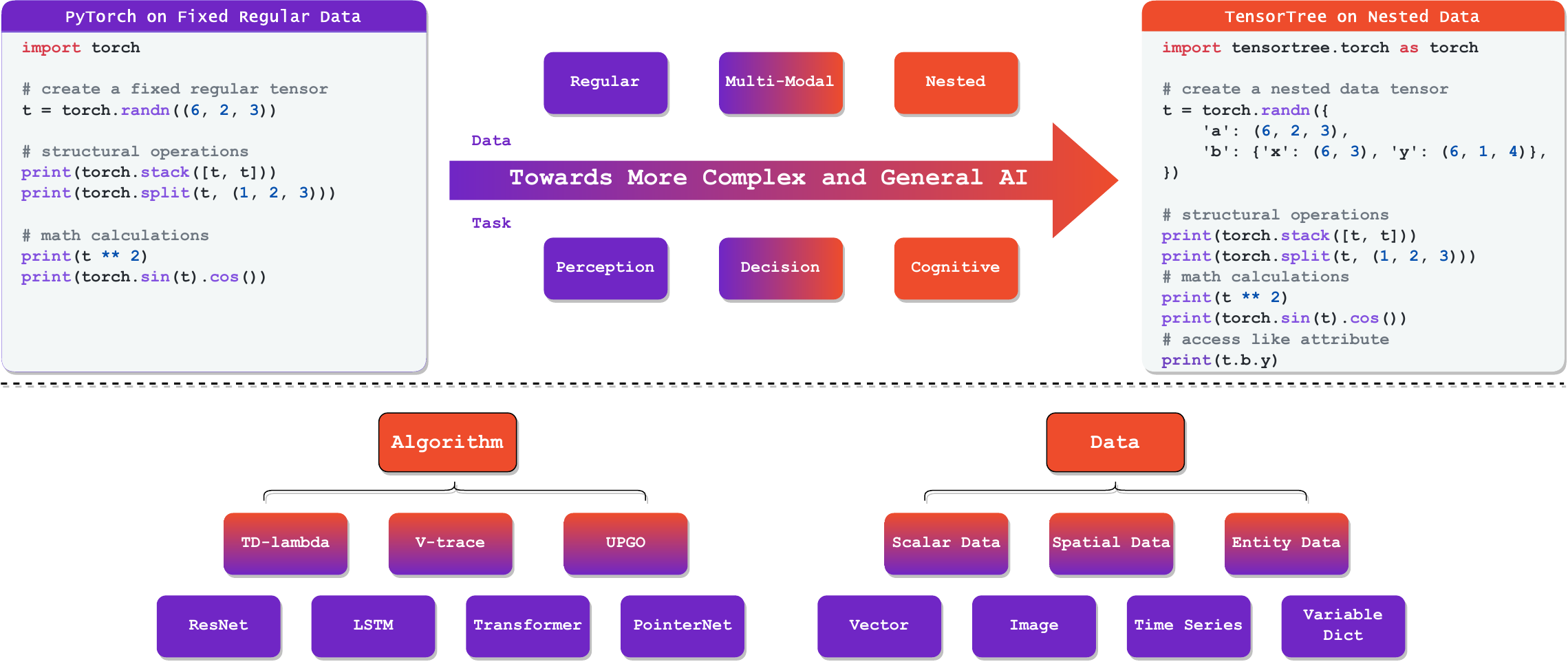}
    \caption{Up: Comparison and transition between native PyTorch and TreeTensor-augmented PyTorch. Users can handle more complicated AI task on nested using TreeTensor, with the consistent usage experience and zero switch cost. Down: Transition in algorithm and data when progressing towards more complex and general AI system. This example is from related information of AlphaStar \cite{arulkumaran_alphastar_2019} and shows the concrete data structures and algorithms combinations.}
    \label{fig-code-compare}
\end{figure*}

There is a key fact that data modalities in perception AI tasks, such as image, video, and human language, are naturally suitable for Tensor. For example, the shape is same for each channel in an image, each frame in a video or each word token in a sentence. And usually these data barely contains complex tree structure. However, when research interests and industrial demands are gradually beginning to pay more attention to complicated tasks like perception-decision AI \cite{montfort_2009_atari, berner_dota_2019, degrave2022magnetic, reed2022generalist}, and try to solve more complex and general AI problems, more diverse data comes like a tide and no longer has the same attributes as Tensor (e.g. data structure of AlphaStar \cite{arulkumaran_alphastar_2019} shown in bottom right of Figure \ref{fig-code-compare}).
The data complexity, multiple modality and nested structure highly expand together across different AI problems. As a result, the classic Tensor is losing the capability to deal with increasingly complicated scenarios. Besides, this so-called nested data can't fully utilize existing general computational unit and vectorized instruction set. Specifically, deep reinforcement learning (DRL) \cite{sutton2018reinforcement} is one of the most heavily influenced areas caused by nested data. For example, one of the most traditional DRL algorithm: DQN \cite{mnih_dqnnature_2015}, can be implemented on naive environment with 5-10 lines of core code, but will redundantly consumes tenfold in nested data setting.

Some researchers take advantage of the combination of native Python list and dict to handle this problem instead, resulting in heavy roundabout effort organizing data structure suitable for complicated operations. Also, it requires the knowledge of the entire data structure. 
Recently, in order to address this issue, dm-tree \cite{beattie_deepmind_2016}, jax-libtree \cite{bradbury_jax_2018} and torchbeast \cite{kuttler_torchbeast_2019} try to define logical structure of these data, and implement some general interfaces like \textit{mapping, flatten}. Besides, RaggedTensor \cite{abadi_tensorflow_2016} and nestedTensor \cite{noauthor_nestedtensor_2022} are proposed to mainly deal with variable-length data, but its re-implementation of Tensor operations adds to heavy workload and extensibility.

In this paper, we examine the data characteristics of various AI tasks and summarize related operations into two fundamental computation modes: 1) \textit{Apply a unary function to all nodes in a single tree}. 2) \textit{Operate function between two or more trees} (the definition of tree is detailed in Section \ref{subsection-method-definition}). 
Based on these insights, we propose a new nested data container named \textbf{TreeTensor}, which incorporates the benefits of the above-mentioned techniques and is more suitable for nested data and future AI tasks (illustrated in Figure \ref{fig-code-compare}).

Firstly, TreeTensor unifies interfaces of processing nested data instances by expanding the definitions and operations of a tree structure, as a result, one can not only access parallel computing as a regular Tensor, but also represent the underlying logic connections between data fields by utilizing properties of tree.
Secondly, the resistance of consistently handling variable-length data are separated into two categories: structure mismatch and shape mismatch. The former can be addressed by our pre-defined four policies (Strict, Inner, Outer, Left), and the latter can be translated into a sub-problem that we can specialize multiple backends to figure it out, such as the existing method like nestedtensor approach or our group padding mechanism.
Thirdly, we also design inheritable constraint mechanism to formulate the behaviour of TreeTensor, which solves the most important overhead when using native Python list and dict. With our proposed multiple constraints, the usability of TreeTensor can be further improved.
Supported by the aforementioned design concepts and various implementation utilities, TreeTensor can improve programming usability and parallel efficiency in many deep reinforcement learning scenarios, while retaining enough extensibility to adapt any new functions and libraries with almost zero cost. 
To show the practical use, we first demonstrate a series code examples about illustration of programming extensibility, mismatch policies and multiple constraints. More importantly, we utilize TreeTensor to boost many practical AI algorithms and applications, especially optimizing some code in AlphaStar \cite{arulkumaran_alphastar_2019}, demonstrating that TreeTensor does make the programming experience much easier.
Besides, comprehensive benchmark results about the typical operations indicate that our implementations are as well as or even better than similar libraries without any overhead.

%% file: sections/2.related_works.tex
\section{Related Works}
\label{section-related}

%
%

\begin{figure*}[htb]
    \centering
    \includegraphics[width=0.90\textwidth]{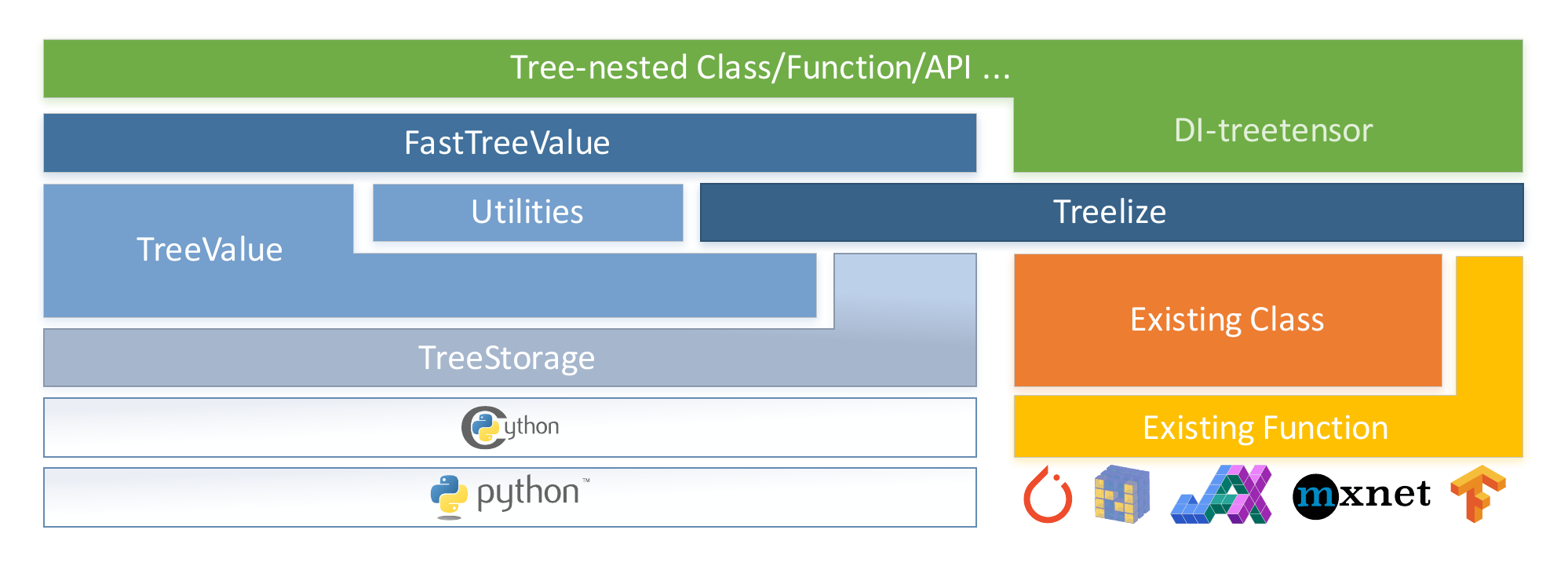}
    \caption{TreeTensor's Overview. The names appearing in this figure are the names appearing in the engineering design. TreeTensor is a Python-based application that using Cython for speedup. \emph{TreeStorage} is the lowest layer, and it is used to encapsulate the tree structure. \emph{TreeValue} class, tree calculating utilities, and the \emph{treelize} tool can then be created. Current library functions, classes, and APIs can be enhanced and merged with the \emph{FastTreeValue} class and additional integrated functions, which can totally inherit the original features.}
    \label{fig-treevalue-conception}
\end{figure*}

Previous works can be roughly divided into two categories: Firstly, dm-tree \cite{beattie_deepmind_2016}, jax-libtree \cite{bradbury_jax_2018} and Torchbeast \cite{kuttler_torchbeast_2019} focus on the logical relationship and structure, and implement some general interfaces like \textit{mapping, flatten} with different system programming tools. Besides, Tianshou Batch \cite{weng_tianshou_2021} is a simple yet effective example in concrete DRL algorithms, but its non-optimized implementation leads to efficiency loss. Secondly, RaggedTensor \cite{abadi_tensorflow_2016} and nestedtensor \cite{noauthor_nestedtensor_2022} pay more attention to variable-length data, in which case similar data might have different lengths. However, these methods have to refactor most mechanisms about Tensor, including storage and internal CUDA kernel, which needs tons of workload and is hard to extend. Compared to previous design, the above mentioned libraries just focus on some specific functions and scenarios, and our treetensor can work on arbitrary nested data operations with few programming and runtime cost. Besides, due to enough scalability, treetensor can also be combined with some libraries like nestedtensor to further improve performance at a special area.


%% file: sections/3.treevalue.tex
\section{TreeTensor}
\label{section-method}


The overview of TreeTensor design is shown in Figure \ref{fig-treevalue-conception}, more details are shown in Appendix A. Then we introduce \textbf{TreeTensor} as follows: Firstly, we start from analyzing the data trends of current AI system. In Section \ref{subsection-method-definition}, we will define the tree-nested structure and related notations. After that, the key feature named \textbf{treelize} is illustrated in Section \ref{subsection-method-treelize}, which is designed to fundamentally improve scalability. Besides, the mismatch polices and the property-based constraint system for enhancing functional component will be described in Section \ref{subsection-method-constraint} and Section \ref{subsection-method-mismatch} respectively. At last, we add an additional part for performance optimization of TreeTensor in Section \ref{subsection-method-optimization}.


\subsection{Data Trends in Perception-Decision AI System}
\label{trend}
In classic perception AI problems like face recognition \cite{deng_arcface_2019} and machine translation \cite{devlin_bert_2018}, regular tensor are the most common data formats, but nested data is becoming increasingly important, such multi-modal structured observation in AlphaStar (shown in down right of Figure \ref{fig-code-compare}), including 2D matrix feature layer (\textit{i.e.}, spatial observation), scalar and vector observation, variable-length entity tensor, human statistics vector $z$ occasionally needed and so on. Except for observation, there are corresponding nested examples for other elements in DRL,  e.g.,  action \cite{milani_minerl_2020}, reward \cite{berner_dota_2019} and so on.
Simultaneously, the combination of different algorithms (shown in down left of Figure \ref{fig-code-compare}) might result in complicated data structure since distinct algorithm modules usually create various attributes and must maintain them throughout different execution components, increasing complexity further.

\subsection{Tree-Nested Structure Definition}
\label{subsection-method-definition}

\begin{figure}[htb]
    \centering
    \includegraphics[width=0.6\linewidth]{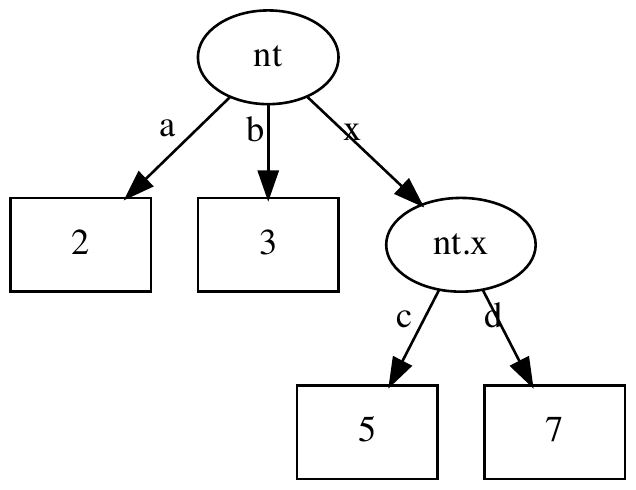}
    \caption{Tree Nodes $n^t$ and $n^t_x$'s Structure. Square nodes indicate value nodes, while circular nodes represent tree nodes.}
    \label{fig-tree-node-example-n}
\end{figure}

Node is the first core concepts in our work. There are two different kinds of nodes: value nodes and tree nodes. The tree node $n^t$ indicates a sub tree, while the value node $n^v$ represents a specific value. 
Since we can denote the name of each sub tree as key $k$, then we can utilize the pair notation $\left\langle k, n \right\rangle$ to describe the entire tree node as a set of $m$ pairs, as is demonstrated in Expression \ref{formula:tree_node}.
\begin{align}
    n^v = \left\langle v \right\rangle, n^t = \left\{ \left\langle k_1, n_1\right\rangle, \left\langle k_2, n_2\right\rangle, \cdots, \left\langle k_m, n_m\right\rangle \right\} \label{formula:tree_node}
\end{align}
For example, the Expression \ref{formula:n_example} represents a tree node which has a total of 5 key-node pairs with 4 value nodes $n^v_a$, $n^v_b$, $n^v_{x,c}$, and $n^v_{x,d}$. Also, we visualize its structure in Figure \ref{fig-tree-node-example-n}.
\begin{align}
    n^v_a &= \left\langle 2 \right\rangle, n^v_b = \left\langle 3 \right\rangle, n^v_{x,c} = \left\langle 5 \right\rangle, n^v_{x,d} = \left\langle 7 \right\rangle \notag \\
    n^t_x &= \left\{ \left\langle\text{'c'}, n^v_{x,c}\right\rangle, \left\langle\text{'d'}, n^v_{x,d}\right\rangle \right\} \label{formula:n_x_example} \\
    n^t &= \left\{ \left\langle\text{'a'}, n^v_a\right\rangle, \left\langle\text{'b'}, n^v_b\right\rangle, \left\langle\text{'x'}, n^t_x\right\rangle \right\} \label{formula:n_example}
\end{align}
In summary, the whole data structure, as shown in Figure \ref{fig-tree-node-example-n}, is composed of a series of value nodes containing data and a tree structure that organizes the relationships between data. It should only have one tree node as the entire tree's root. We call these tree-nested structure as TreeTensor.

\subsection{Treelize}
\label{subsection-method-treelize}

In order to support complex and diverse tensor operations on tree structures, we design the \textit{treelize} method to extend existing function to its tree-nested counterpart easily. Treelize for unary and multivariate operations are defined in Section \ref{subsubsection-method-treelize-unary} and Section \ref{subsubsection-method-treelize-multi} respectively. And we introduce how to apply treelize to existing API in Section \ref{subsubsection-method-treelize-extension}.

\subsubsection{Apply a Unary Function to All Nodes in a Single TreeTensor}
\label{subsubsection-method-treelize-unary}

Based on TreeTensor data structure established in Section \ref{subsection-method-definition}, we first introduce unary functions to all nodes in a single tree, which can represented in Expression \ref{formula:unary_operation}. The unary functions are applied to each value of the TreeTensor, producing a new tree with the same tree structure.
\begin{align}
    y = f\left(x\right) \label{formula:unary_operation}
\end{align}

\begin{figure}[htb]
    \centering
    \subfigure[TreeTensor $n_1$'s Structure.]{
        \centering
        \label{fig-all-operate-example-n1}
        \includegraphics[width=0.45\linewidth]{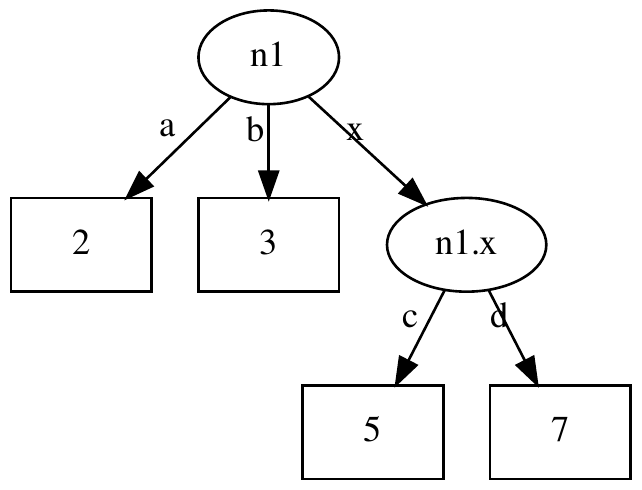}
    }
    \subfigure[Obtained TreeTensor $n^{\prime}_1$'s Structure.]{
        \centering
        \label{fig-all-operate-example-nx-1}
        \includegraphics[width=0.45\linewidth]{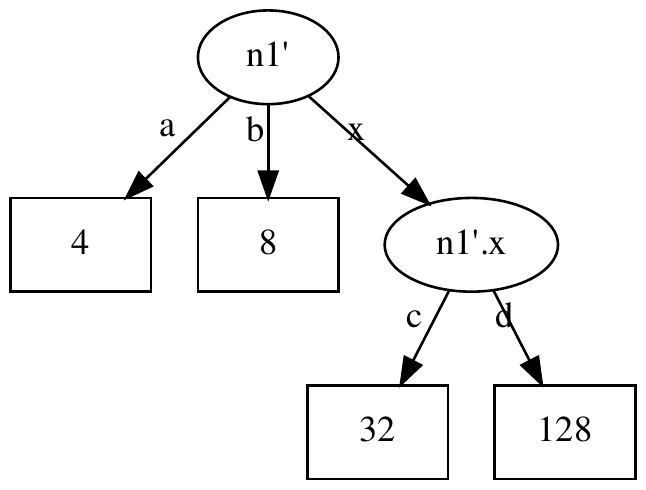}
    }
    \caption{Operate Function $p$ on All the Nodes Of Tree $n_1$. The structure of $n^{\prime}_1$ is identical to that of $n_1$, but its values are handled by function $p$.}
    \label{fig-all-operate-example}    
\end{figure}

When the unary function operates on a value node $n^v$, the result is a new value node named $n^{v \prime}$, shown in the Expression \ref{formula:unary_nx_v}. This procedure can be defined as a function $F$ from $n^v$ to $n^{v \prime}$. Similarly, the function $F$ for tree node in the Expression \ref{formula:unary_nx_t} can be defined based on above definitions. 
For example, in Figure \ref{fig-all-operate-example-n1}, there is a TreeTensor named $n_1$. We can define a unary function $p\left(x\right) = 2 ^ x$ and apply it over the entire $n_1$, with the output $n^{\prime}_1$ described in Figure \ref{fig-all-operate-example-nx-1}.
\begin{align}
    F\left(n^v\right) &= n^{v \prime} = \left\langle f\left( v \right)\right\rangle \label{formula:unary_nx_v} \\
    F\left(n^t\right) &= n^{t \prime} = \left\{ \left\langle k_1, F\left(n_1\right) \right\rangle, \cdots, \left\langle k_m, F\left(n_m\right) \right\rangle \right\} \label{formula:unary_nx_t}
\end{align}

\subsubsection{Operate Function Among Two or More TreeTensor}
\label{subsubsection-method-treelize-multi}

Furthermore, a binary, ternary, or multivariate function can be applied to multiple trees in a similar way. A structure shown in the Expression \ref{formula:multi_operation} can be used to define this type of function\footnote{When $c = 1$ , the unary function stated in Expression \ref{formula:unary_operation} is essentially a special case of this.}.
\begin{align}
    y = f\left( x_1, x_2, \cdots, x_c \right), c \geq 1 \label{formula:multi_operation}
\end{align}

When this multivariate function is applied to numerous value nodes $n^v_1, n^v_2, \cdots, n^v_c$, the result is a new value node named $F\left(n^v_1, n^v_2, \cdots, n^v_c\right)$ (Expression \ref{formula:multi_nx_v}). We can also define this process as function $F$. Based on these definitions, the function $F$ for tree node can be reprensented in the Expression \ref{formula:multi_nx_t} as follows:
\begin{align}
    F\left(n^v_1, n^v_2, \cdots, n^v_c\right) =& \left\langle f\left( v_1, v_2, \cdots, v_c \right) \right\rangle \label{formula:multi_nx_v} \\
    F\left(n_1, n_2, \cdots, n_c\right) =& \{ \left\langle k_1, F\left( n_{1, 1}, \cdots, n_{1, c} \right) \right\rangle,  \notag \\
    & \cdots, \notag \\
    & \left\langle k_m, F\left( n_{m,1}, \cdots, n_{m,c} \right) \right\rangle  \} \label{formula:multi_nx_t}
\end{align}

\begin{figure}[htb]
    \centering
    \subfigure[$n_1$'s Structure.]{
        \label{fig-multi-operate-example-n1}
        \includegraphics[width=0.22\textwidth]{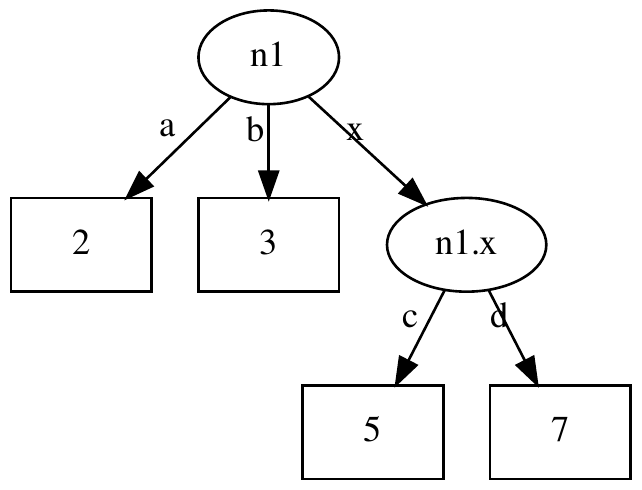}
    }
    \subfigure[$n_2$'s Structure.]{
        \label{fig-multi-operate-example-n2}
        \includegraphics[width=0.22\textwidth]{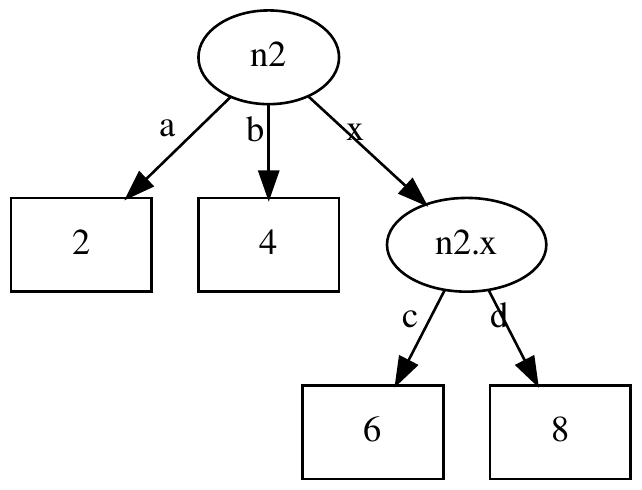}
    }
    \subfigure[$n_3$'s Structure.]{
        \label{fig-multi-operate-example-n3}
        \includegraphics[width=0.22\textwidth]{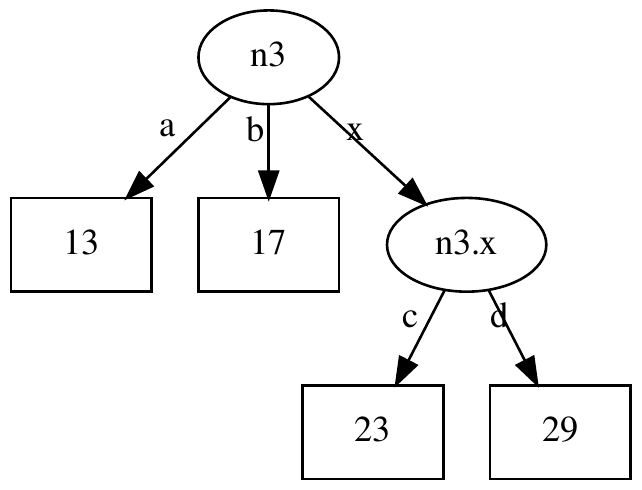}
    }
    \subfigure[Obtained TreeTensor $n^{\prime}$'s Structure.]{
        \label{fig-multi-operate-example-nx}
        \includegraphics[width=0.23\textwidth]{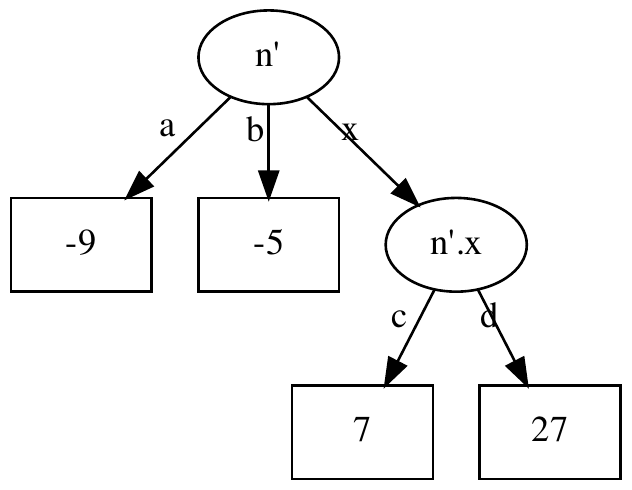}
    }
    \caption{Operate Function $h$ Among Trees $n_1$, $n_2$ and $n_3$. The structure of $n^{\prime}$ is identical to the original trees, but its values are handled by function $h$.}
    \label{fig-multi-operate-example}
\end{figure}

For Instance, Figure \ref{fig-multi-operate-example} shows three TreeTensor labeled $n_1$, $n_2$, and $n_3$. We can create a ternary function $h\left(x, y, z\right) = x \cdot y - z$ and apply it on three TreeTensors, yielding the final result $n^{\prime}$ as illustrated in Figure \ref{fig-multi-operate-example-nx}.

\subsubsection{Extending the Functionality of Existing API}
\label{subsubsection-method-treelize-extension}

In practice, the most critical issue of our work is how to use treelize in some existing libraries and frameworks which are already stable. It is impractical and inconvenient to make many changes or assumptions about its underlying design. As a result, a non-intrusive approach is required so that the above operating features can be applied to any existing libraries or frameworks.

To address this issue, we extend the definitions of Expression \ref{formula:multi_nx_v} and Expression \ref{formula:multi_nx_t} mentioned above to have the native value in this operation and to have the same status as the value node, as shown in the Expression \ref{formula:multi_f_extend} \footnote{In the Expression \ref{formula:multi_f_extend}, $f$ means the original function, while $F$ means the extended function which supports calculation based on TreeTensor.}.
\begin{align}
    F\left(v_1, v_2, \cdots, v_c\right) &= F\left(n^v_1, n^v_2, \cdots, n^v_c\right) \notag \\
    &= f\left(v_1, v_2, \cdots, v_c\right) \label{formula:multi_f_extend}
\end{align}

As a result, as described in the phrase \ref{formula:delta_f_to_f}, we may define the process from $f$ to $F$ as a function $\Delta_\text{func}$. \emph{Treelize} is the name of the function $\Delta_\text{func}$, which converts the original function $f$ to a tree-supported function $F$.
\begin{align}
    \Delta_\text{func}\left(f\right) &= F \label{formula:delta_f_to_f}
\end{align}

The \emph{treelize} operation can be be extended to classes using the treelize function mentioned above. From a business standpoint, the class structure can be thought of as a key-value pair structure, with the key being the method name and the value being the method itself\footnote{In fact, because the constructor is a special method, this model can also be used to summarize it.}. The Expression \ref{formula:delta_c_to_c} is used to define the \emph{treelize} of class. As a result, $\Delta_\text{func}$ can be applied to a variety of methods, resulting in the creation of a new class. This is known as \emph{class treelize}, and it is expressed as $\Delta_\text{class}$.
\begin{align}
    c &= \left\{ \left\langle p_1, f_1 \right\rangle, \left\langle p_2, f_2 \right\rangle, \cdots, \left\langle p_n, f_n \right\rangle \right\} \\
    C &= \Delta_\text{class}\left( c \right) \notag \\
    &= \left\{ \left\langle p_1, \Delta_\text{func}\left(f_1\right)\right\rangle  , \cdots, \left\langle p_n, \Delta_\text{func}\left( f_n \right) \right\rangle \right\} \label{formula:delta_c_to_c}
\end{align}

The above extension features, combined with a number of Python features such as decorators, make it simple to extend the operation based on tree structure to any existing interfaces while maintaining the original operation properties and making them suitable for tree-nested arguments.

\subsection{Policy When Structure Mismatch}
\label{subsection-method-mismatch}

In some circumstances, there may be mismatches among nodes especially when doing some multivariate operations among different trees. Mismatches among keys of different trees are the most important reasons. We give four policies to cope with possible cases of tree node key mismatching, as indicated in Table \ref{tab:mode-for-key-mismatching}. Section \ref{subsection-examples-mismatch} and Appendix C show more information and examples.
\begin{table}[htb]
\centering
\begin{tabular}{cccc}
\hline
Policy & \begin{tabular}[c]{@{}c@{}}Allow\\ Mismatch\end{tabular}   & Key Set      & \begin{tabular}[c]{@{}c@{}}Default\\ Value\end{tabular} \\ \hline
Strict & \xmark     & Any(All)     & \xmark        \\
Inner  & \cmark     & Intersection & \xmark        \\
Outer  & \cmark     & Union        & \cmark        \\
Left   & \cmark     & First One    & \cmark        \\ \hline
\end{tabular}
\caption{Four Policies for Dealing With Key Mismatching. All matching keys must correspond absolutely one to one only in the strict mode, which is also the default option. The other three modes each has their own methods for dealing with key mismatch, and some of them require default values.}
\label{tab:mode-for-key-mismatching}
\end{table}


\subsection{Constraint and Feature Expansion}
\label{subsection-method-constraint}

In some practical application scenarios, the values on each leaf node in TreeTensor will satisfy a certain relationship, such as Tensors of type float32, stored in CUDA:0, and have a common prefix on the shapes of each Tensors. Obviously, when more properties are assumed, the computing features it can provide will also be expanded. So we designed the constraint system for TreeTensor.

\subsubsection{Definition of Constraints and Basic Properties}
\label{subsubsection-method-constraint-definition}

For a node $n$ in TreeTensor, we define its constraint $C_n$. The constraint can be used as a judgment function for the node, that is, when $C_n\left(n\right)$ is true, it means that node n satisfies the constraint $C_n$, otherwise it does not satisfy the constraint. On this basis, we call the set of all possible nodes that can satisfy the constraint $C$ as $D_C$, so it has the following operational properties:
\begin{itemize}
    \item When all nodes that satisfy $C_2$ can satisfy $C_1$, we say $C_1$ covers $C_2$, denoted as $C_1 \sqsupseteq C_2$, as shown in Expression \ref{formula:constraint_cover}.
    \item When $C_1$ covers $C_2$ and $C_2$ covers $C_1$, we say $C_1$ equals to $C_2$, denoted as $C_1 = C_2$, as shown in Expression \ref{formula:constraint_equal}.
    \item When any possible nodes can satisfy $C$, we say $C$ is an empty constraint, denoted as $C = C_\emptyset$, as shown in Expression \ref{formula:constraint_empty}.
    \item When $C_3$ is satisfied if and only if $C_1$ and $C_2$ are both satisfied, we say $C_3$ equals to $C_1$ plus $C_2$, denoted as $C_3 = C_1 + C_2$, as shown in Expression \ref{formula:constraint_add} and Expression \ref{formula:constraint_sum}.
\end{itemize}
\begin{align}
    D_C &= \left\{ n | C\left(n\right) \right\} \label{formula:data_of_constraint} \\
    C_1 \sqsupseteq C_2 &\iff D_{C_1} \subseteq D_{C_2} \label{formula:constraint_cover} \\
    C_1 = C_2 &\iff C_1 \sqsupseteq C_2 \land C_2 \sqsubseteq C_1 \label{formula:constraint_equal} \\
    C = C_\emptyset &\iff \forall n, n \in D_C \label{formula:constraint_empty} \\
    C_3 = C_1 + C_2 &\iff D_{C_3} = D_{C_1} \cap D_{C_2} \label{formula:constraint_add} \\
    C = \sum\limits_{i=1}^{n}{C_i} &\iff D_C = \bigcap\limits_{i=1}^{n}{D_{C_i}} \label{formula:constraint_sum} 
\end{align}

\subsubsection{Inheritance of Constraints}
\label{subsubsection-method-constraint-inherit}



For the practical application of TreeTensor, it is not difficult to find that constraints need to act on two types of situations, one is the constraints that act on all value nodes in the subtree, such as "is float32 type", "stored in CUDA:0", etc., and the other is the constraints that act on multiple subtrees and child nodes, such as "the Tensor's shape of the 'anchor' value node and the 'positive'/'negative' value node is consistent and can be used for metric learning computing". To this end, we define two types of constraints, inheritance constraints and non-inheritance constraints.

For inheritance constraint, it can be define as Expression \ref{formula:constraint_inherit}, containing one check function on value, denoted as $p_I$. For value node, inheritance constraint $C_I$ is satisfied if and only if $P^I\left(v\right)$ is true, as shown in Expression \ref{formula:constraint_inherit_value}. For tree node, $C_I$ is satisfied if and only if all $n^t$'s child node satisfy $C_I$, as shown in Expression \ref{formula:constraint_inherit_tree}. This kind of constraint can be used to modeling constraints 'is float32 type' and 'stored in CUDA:0'. On this basis, due to the property in Expression \ref{formula:constraint_inherit_tree}, we define the inherit function (denoted as $\Psi$, will be used in Section \ref{subsubsection-method-constraint-tree}), and the result of $\Psi\left( C^I \right)$ (named as 'inherited constraint') should still be $C_I$.
\begin{align}
    C^I &= \left< p^I \right> \label{formula:constraint_inherit} \\
    C^I\left( n^v \right) &\iff p^I\left(v\right) \label{formula:constraint_inherit_value} \\
    C^I\left( n^t \right) &\iff \forall \left< k_i, n_i \right> \in n^t, C^I\left( n_i \right) \label{formula:constraint_inherit_tree} \\
    \Psi\left( C^I \right) &= C^I \label{formula:inherit_constraint_inherit}
\end{align}
For non-inheritance constraint, it is defined as Expression \ref{formula:constraint_non_inherit}, containing one check function on node, denoted as $p^N$. Constraint $C^N$ is satisfied by node $n$ if and only if $p^N\left( n \right)$ is true, as shown in Expression \ref{formula:constraint_non_inherit_node}. Obviously, because the non-inheritance constraints are fixed on specific nodes, so its inherited  constraint should be empty constraint, as shown in Expression \ref{formula:inherit_constrainnt_non_inherit}.
\begin{align}
    C^N &= \left< p^N \right> \label{formula:constraint_non_inherit} \\
    C^N\left( n \right) &\iff p^N\left( n \right) \label{formula:constraint_non_inherit_node} \\
    \Psi\left( C^N \right) &= C_\emptyset \label{formula:inherit_constrainnt_non_inherit}
\end{align}
In a further case, there will be a class of non-inheritance constraints that impose constraints on multiple different child nodes under a subtree, such as $a \cdot \left(b_x + b_y\right) > 0$ \footnote{$a$, $b_x$ and $b_y$ can be seen as value of value nodes $n^v_a$, $n^v_{b,x}$ and $n^v_{b,y}$.}. For such a case, the multivariate function can be split in the form of a partial function. For example, the above constraint should be split into three checking functions of $n^v_a$, $n^v_{b,x}$ and $n^v_{b,y}$ as variables, and they should be constructed as three non-inheritance constraints on the corresponding nodes.

\subsubsection{Constraint Tree And Validation}
\label{subsubsection-method-constraint-tree}



Based on the definition and properties of constraints, as well as the inheritances, we can construct a constraint tree for TreeTensor. It is worth noting that the constraint tree and the TreeTensor are a one-to-one combination, that is, the constraint tree is a necessary part of the TreeTensor. 

\begin{figure}[htb]
    \centering
    \subfigure[TreeTensor $n$.]{
        \label{fig-constraint-tree-example}
        \includegraphics[width=0.47\linewidth]{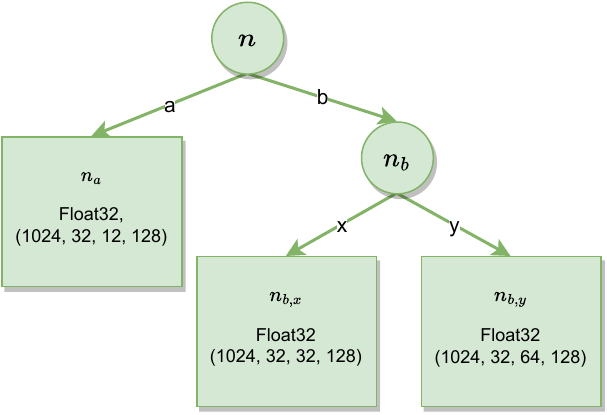}
    }
    \subfigure[Original Constraint Tree.]{
        \label{fig-constraint-init-example}
        \includegraphics[width=0.47\linewidth]{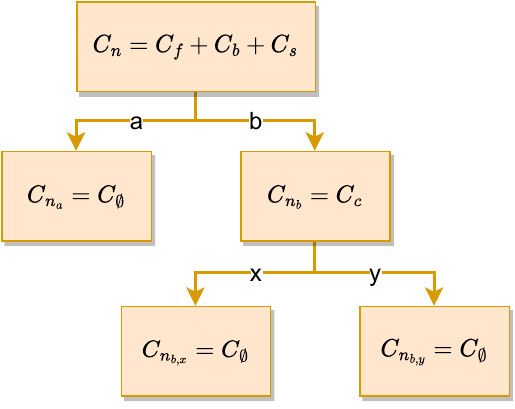}
    }
    \subfigure[TreeTensor $n$ With The Final Constraint Tree.]{
        \label{fig-constraint-final-example}
        \includegraphics[width=0.97\linewidth]{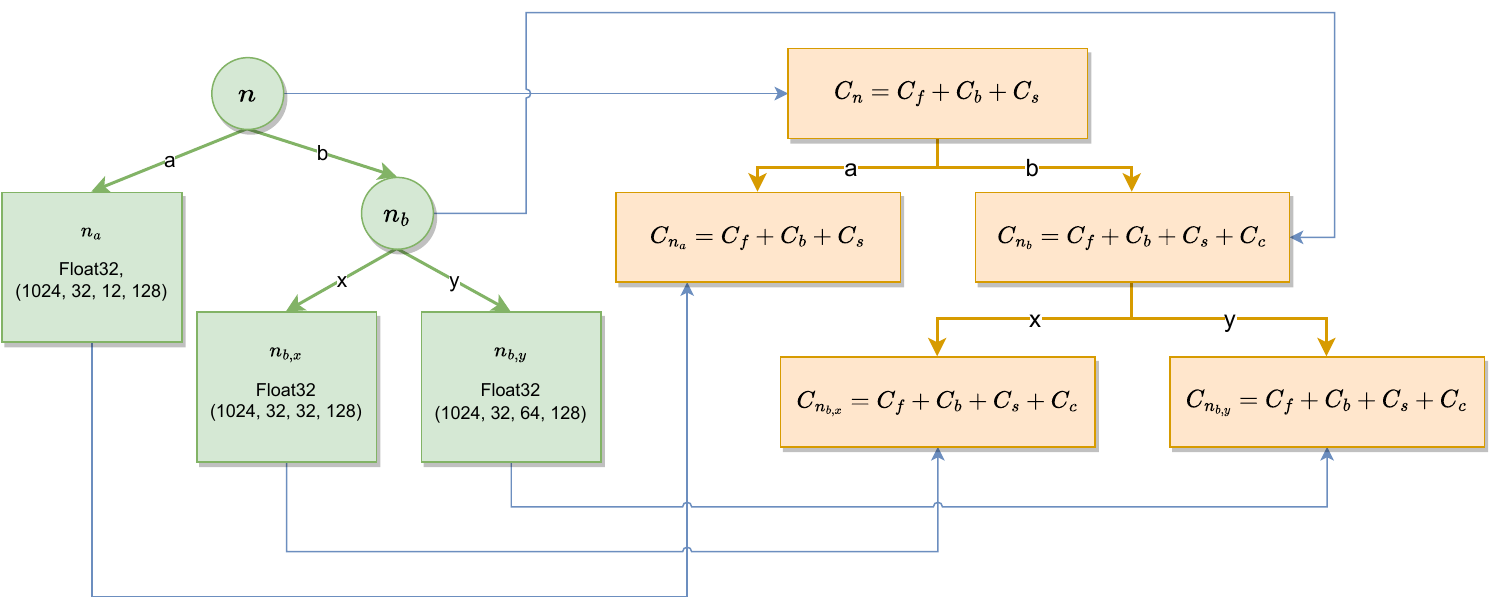}
    }
    \caption{Example and Result of Constraint Tree's Construction.}
    \label{fig-constraint-example}
\end{figure}

For the sake of illustration, we will describe the construction and validation of the constraint tree with examples. Consider the TreeTensor shown in Figure \ref{fig-constraint-tree-example}. The internal value nodes are 4-dimensional tensors in the format of float32, which are used for batch metric learning. The first two dimensions of these Tensors are 1024 and 32, which constitute multiple batches of sample sets. The dimensions count of a single sample is 128. The value node $n_a$ contains anchor samples of metric learning, while the value node $n_{b,x}$ and $n_ {b,y}$ contains positive and negative samples respectively, and their sample sizes are not less than 24. For this requirement, we can define four inheritance constraints as shown from Expression \ref{formula:constraint_cf}-\ref{formula:constraint_cc}. From this, we can start the construction of a TreeTensor. First, organize the data in the way shown in Figure \ref{fig-constraint-tree-example}, and then place corresponding constraints on the corresponding positions of each node\footnote{Unconstrained nodes can be regarded as constrained by empty constraints, denoted as $C_\emptyset$. See Section \ref{subsubsection-method-constraint-definition}.}, as shown in Figure \ref{fig-constraint-init-example}. On this basis, distribute the constraint tree. For any non root node constraint $C'$ and its parent node constraint $C$, execute $C' \leftarrow C' + \Psi\left(C\right)$ until there is no change in the entire constraint tree. Combine each tree node or value node with the constraint tree node at the corresponding position, as shown in Figure \ref{fig-constraint-final-example}, to complete the construction of a TreeTensor with constraints.
\begin{align}
    C_f &= \left< \text{float32, 4 dims, tensor} \right> \label{formula:constraint_cf} \\
    C_b &= \left< \text{1st, 2nd dims are 1024, 32} \right> \label{formula:constraint_cb} \\
    C_s &= \left< \text{4th dim is 128} \right> \label{formula:constraint_cs} \\
    C_c &= \left< \text{3rd dim no less than 24} \right> \label{formula:constraint_cc}
\end{align}

After the constraint tree is constructed, it will not be changed through the entire life cycle of the TreeTensor. For each tree node, its corresponding constraint tree node will be accessible. Therefore, when adding, deleting, and modifying any position of the tree, the corresponding constraint check can be performed when modifying the parent tree node to ensure that all constraints in the whole life cycle of the TreeTensor will always be satisfied. In addition, considering that objects such as tensor are mutable \footnote{Tensor's inner value can be modified after calling some methods without changing the pointer (e.g. sin\_, sigmoid\_, and other methods ends with underline), so that its parent tree node will be not able to be informed when this happened.}, we additionally provide a manual constraint validation method to ensure that the TreeTensors participating operations satisfy the required constraints.

\subsection{Performance Optimization}
\label{subsection-method-optimization}


TreeTensor is written in the Python programming language, with the main components written in Cython \cite{behnel_cython_2011} for speedup. Its core idea is to write code in a syntax that is similar to Python's, then compile it into a static library that can be used in Python. To reduce the time cost of dynamic type determination in Python, one crucial component of optimization is to explicitly define data types and preset memory. In addition, inlining tiny logic blocks which are often called, directly using native data types in internal modules, and removing as many superfluous encapsulation layers as feasible would all help to speed up the procedure. This can be seen more clearly in Cython than CPython.

Not only that, the scalable architecture of TreeTensor also has good compatibility with existing optimization techniques (such as PyTorch's cuda stream), see Appendix J for details.

%% file: sections/4.examples.tex
\section{Examples and Experiments}
\label{section-examples}


\subsection{TreeTensor's Feature}
\label{subsection-examples-treelize}

\subsubsection{What Can Treelize Do}
\label{subsubsection-examples-treelize-extension}


The treelize operation mentioned in Section \ref{subsection-method-treelize} is the core feature of TreeTensor, which can extend various operation functions for ordinary objects to TreeTensor. In Python, all kinds of operations, including ordinary functions, instance methods, operators\footnote{In Python, operators are essentially syntactic sugar based on magic methods, see \cite{python_magic_method}.}, properties, have equivalent form based on ordinary functions, as shown in Table \ref{tab-treelize-equivalents}. And through the treelization of the ordinary functions, all operations in Python can be extended to the tree and supported by TreeTensor.

\begin{table*}[]
\centering
\begin{tabular}{cccc}
                                                & Expression                                & Equivalent Form                                               & \textbf{Function To Be Treelized} \\ \hline
\multicolumn{1}{c|}{Function}                   & \multicolumn{1}{c|}{torch.sigmoid(t)}     & \multicolumn{1}{c|}{torch.sigmoid(t)}                         & torch.sigmoid                     \\ \hline
\multicolumn{1}{c|}{\multirow{3}{*}{Method}}    & \multicolumn{1}{c|}{t.sigmoid()}          & \multicolumn{1}{c|}{Tensor.sigmoid(t)}                        & Tensor.sigmoid                    \\ \cline{2-4} 
\multicolumn{1}{c|}{}                           & \multicolumn{1}{c|}{t.split(3)}           & \multicolumn{1}{c|}{Tensor.split(t, 3)}                       & Tensor.split                      \\ \cline{2-4} 
\multicolumn{1}{c|}{}                           & \multicolumn{1}{c|}{t1.isclose(t2, 1e-5)} & \multicolumn{1}{c|}{Tensor.isclose(t1, t2, 1e-5)}             & Tensor.isclose                    \\ \hline
\multicolumn{1}{c|}{\multirow{4}{*}{Operators}} & \multicolumn{1}{c|}{t1 + t2}              & \multicolumn{1}{c|}{Tensor.\_\_add\_\_(t1, t2)}               & Tensor.\_\_add\_\_                \\ \cline{2-4} 
\multicolumn{1}{c|}{}                           & \multicolumn{1}{c|}{t1 @ t2}              & \multicolumn{1}{c|}{Tensor.\_\_matmul\_\_(t1, t2)}            & Tensor.\_\_matmul\_\_             \\ \cline{2-4} 
\multicolumn{1}{c|}{}                           & \multicolumn{1}{c|}{t1{[}2:-1{]}}         & \multicolumn{1}{c|}{Tensor.\_\_getitem\_\_(t1, slice(2, -1))} & Tensor.\_\_getitem\_\_            \\ \cline{2-4} 
\multicolumn{1}{c|}{}                           & \multicolumn{1}{c|}{t.anyvalue}           & \multicolumn{1}{c|}{Tensor.\_\_getattr\_\_(t1, 'anyvalue')}   & Tensor.\_\_getattr\_\_            \\ \hline
\multicolumn{1}{c|}{\multirow{2}{*}{Property}}  & \multicolumn{1}{c|}{t.shape}              & \multicolumn{1}{c|}{Tensor.shape.\_\_get\_\_(t)}              & Tensor.shape.\_\_get\_\_          \\ \cline{2-4} 
\multicolumn{1}{c|}{}                           & \multicolumn{1}{c|}{t.T}                  & \multicolumn{1}{c|}{Tensor.T.\_\_get\_\_(t)}                  & Tensor.T.\_\_get\_\_              \\ \hline
\end{tabular}
\caption{Functions, Methods, Operators and Properties in Python. All of these operation types have equivalents that use only simple functions, which are able to be treelized. }
\label{tab-treelize-equivalents}
\end{table*}

\subsubsection{TreeTensor on PyTorch}
\label{subsubsection-examples-treelize-torch}

Based on common operations on tree structures and Tensors, we provide some additional utilities to process them. The detailed setting and visualization results can be found in Appendix E and F.

\paragraph{Functional Utilities.} 
Including mapping, mask, filter and reduce methods. Mapping operation is similar to the unary operation defined in Section \ref{subsubsection-method-treelize-unary}. Mask, filter is used for picking up some of the nodes in one TreeTensor. Reduce (also named fold) is a recursive process in functional programming \cite{hughes_why_1989} that uses a binary function and an initial value.

\begin{figure}[htb]
    \centering
    \includegraphics[width=0.9\linewidth]{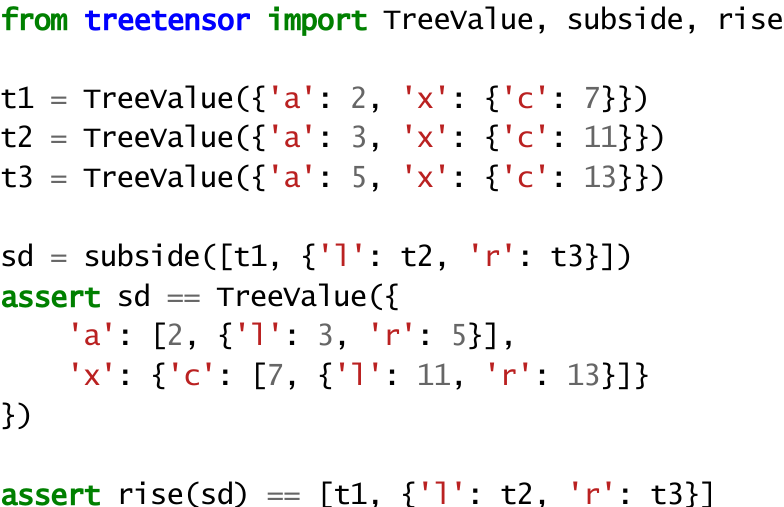}
    \caption{Example of Subside and Rise Operation. The list and dict structures that are above the TreeTensor before subside are subsided to the value nodes. The rise operation is the inverse of the subside, and it returns the list and dict structures to TreeTensor's higher layer.}
    \label{lst:subside-rise-example}
    \vspace{-15pt}
\end{figure}

\paragraph{Structural Utilities.} 
Multiple TreeTensors may be stored under multi-layer data structures (such as nested lists, dicts, tuples, and so on) in Python's specific application, and the values of all value nodes under a single TreeTensor may have similar structure. As a result, we provide the \emph{subside} tool for sinking the external data structure into each value node, and the \emph{rise} tool for rising the common internal data structure from each value node to the top level. \emph{Subside} and \emph{rise} are inverse operations of each other. A sample code is shown in Figure \ref{lst:subside-rise-example}.

\subsubsection{Treelize on Other Libraries}
\label{subsubsection-examples-treelize-library}

Except for PyTorch, practically all functions, classes, and modules can be extended by using \emph{FastTreeValue} and \emph{treelize} operations, which will result in a high level of simplicity of use and a reduction in programming difficulties. This extension also includes Numpy, Figure \ref{lst:numpy-example} shows a simple Numpy extension example. The more detailed examples can be found in in Appendix G, along with the relevant explanation.
\begin{figure}[htb]
    \centering
    \includegraphics[width=0.9\linewidth]{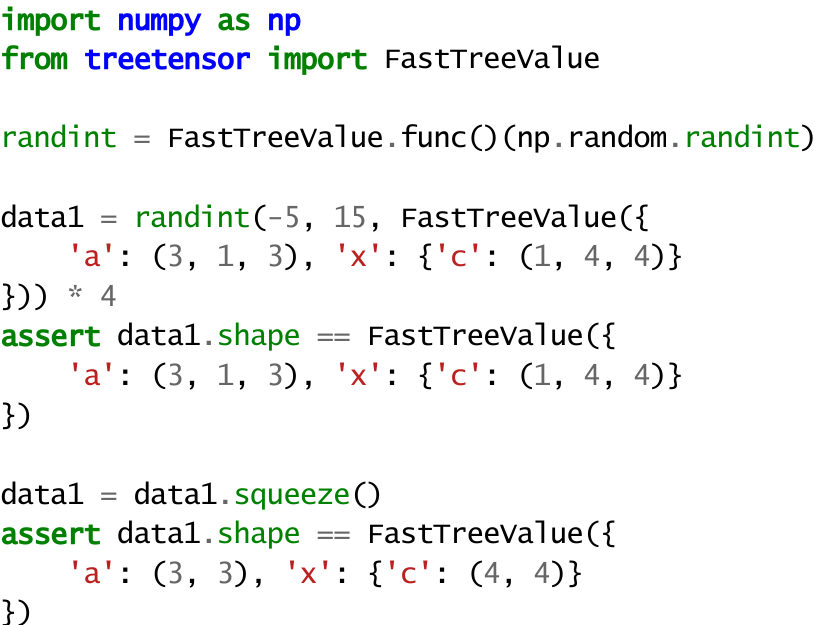}
    \caption{Numpy Extension Example. It is able to generate a tree of ndarrays after treelizing the function \emph{randint}. After Numpy objects being put into \emph{FastTreeValue}, attributes like \emph{shape} and methods like \emph{squeeze} can be accessed and called.}
    \label{lst:numpy-example}
        \vspace{-15pt}
\end{figure}

Moreover, Scikit-Learn \cite{pedregosa_scikit-learn_2011} can be extended with the similar way, as provided in Listing H, will include an example and any relevant explanations.

\subsection{Reasons for Using Mismatch Policy}
\label{subsection-examples-mismatch}

As is shown in Table \ref{tab:mode-for-key-mismatching}, we design four policies for the cases that multiple treetensor owns different tree structures or data formats. For the most cases, we use the strict policy to ensure the correctness of data structure. However, when we want to construct more complicated AI system, data modality and formats will become more and more diverse, such as stream videos, language sentences and game instructions. Even the data structures will be often dynamically changed during training. And we must transform data samples into regular a batch of data for parallel computation and optimization variance reduction. Therefore, we utilize the other three policies to set principles for the operations of treetensor. Besides, we also design special group padding mechanism to speed up variable-length data with the same tree structure, which is described in Appendix B.

\subsection{Significance of Adding Constraints}
\label{subsection-examples-constraints}

\begin{figure}
    \centering
    \includegraphics[width=0.95\linewidth]{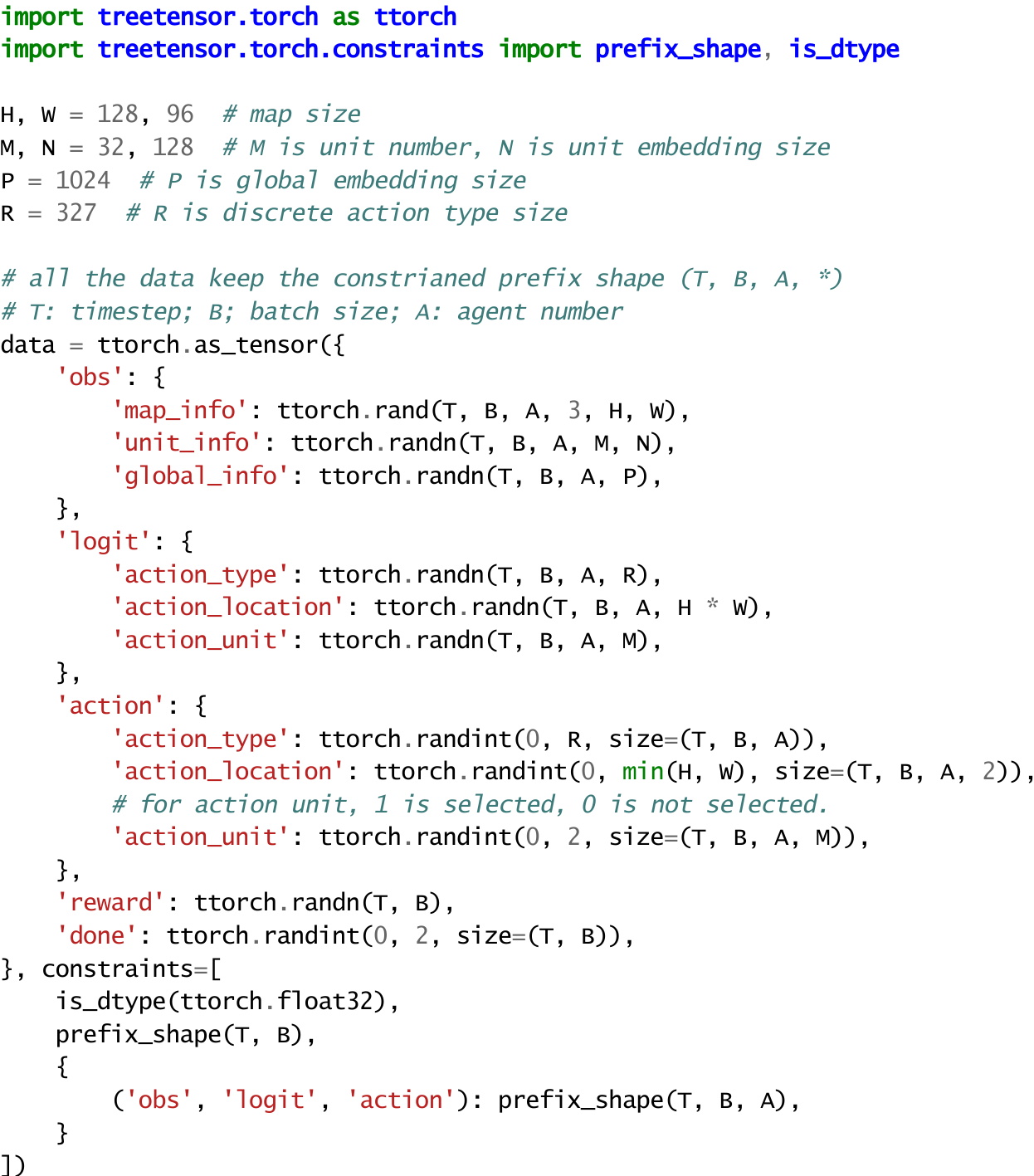}
    \caption{TreeTensor Example For Multi-Agent Reinforcement Learning Training on Long-Horizon Decision.}
    \label{fig-constraint-demo-batch}
    \vspace{-15pt}
\end{figure}

In Section \ref{subsection-method-constraint} we introduce the constraints of TreeTensor. For practical examples, when a TreeTensor can satisfy specific assumptions (specific types, shapes, structures, etc.), it means that more functional extensions can be made on the properties of the original tree. The constraints of TreeTensor provide a way to describe such assumptions and allow developers to build further specialized features for TreeTensor based on the assumption that the construction is complete. As the example below, Figure \ref{fig-constraint-demo-batch} shows a TreeTensor for multi-agent reinforcement learning training, which contains $T$ time steps because of the need to consider the long-term game state, and the data contains $B$ samples. Therefore, for the entire TreeTensor, all internal Tensors should be of type float32, and the shape starts with $T$ and $B$. On this basis, some subtrees store the state data of $A$ agents, so the shapes of tensors inside these subtrees will start with $T$, $B$, and $A$.

\begin{figure}
    \centering
    \includegraphics[width=1.0\linewidth]{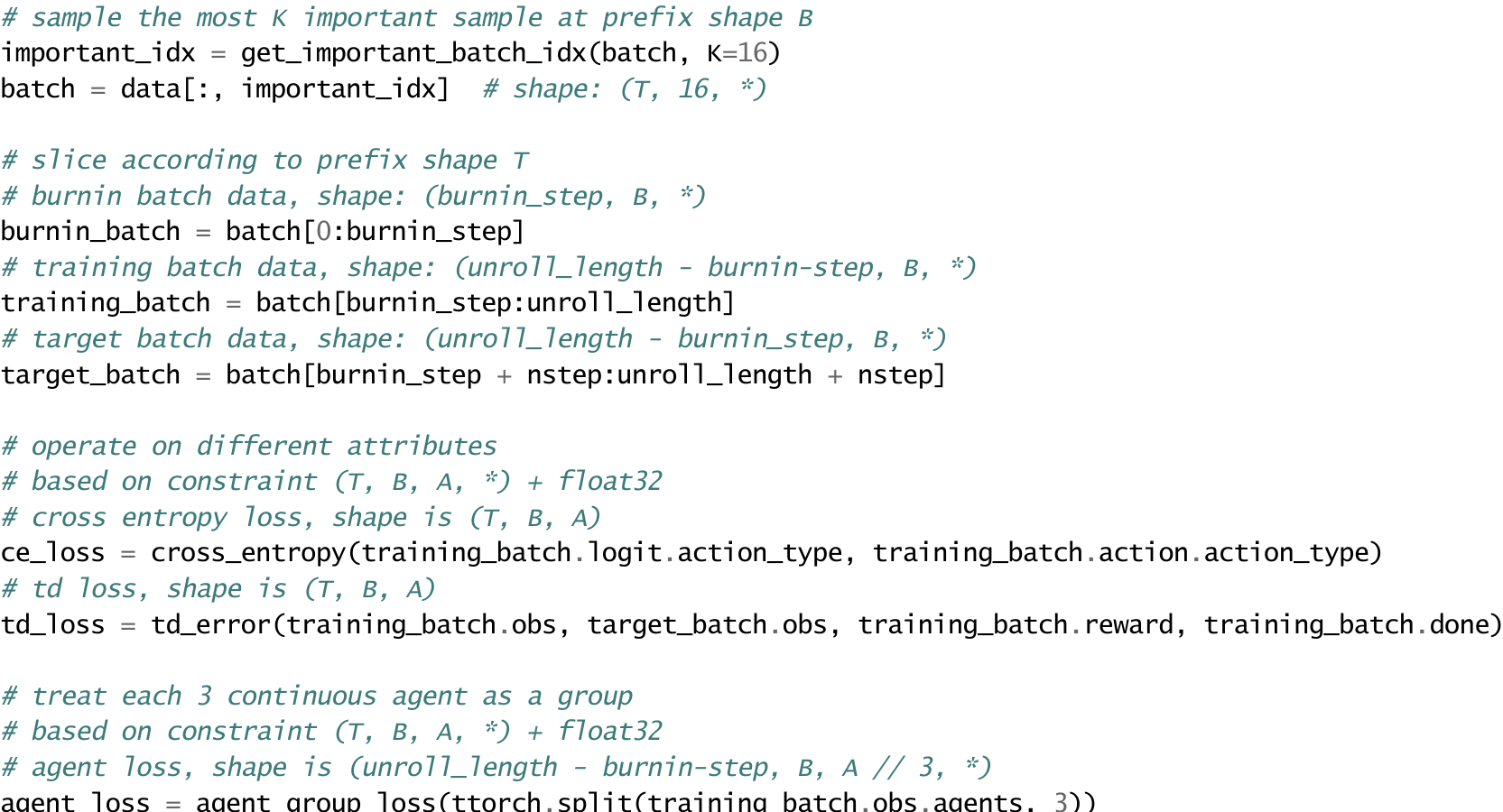}
    \caption{Sampling and Optimization for Data in Figure \ref{fig-constraint-demo-batch}.}
    \label{fig-constraint-demo-run}
        \vspace{-20pt}
\end{figure}

Based on the above constraints, we can sample the data in this format, extract the 16 most important samples and form a new batch as shown in Figure \ref{fig-constraint-demo-run}. In the process of several samplings, the constraints in new TreeTensor will be derived according to the original constraints and the operations performed. After that, the data with the same common shape prefix will be able to be directly used to calculate the loss value, and the shape of the operation result will also be $\left(T, B, A\right)$.

\subsection{Experiment on Pratical Usage}
\label{subsection-examples-actual}

\subsubsection{Code Examples}
\label{subsection-show-example}

In this part, we demonstrate the programming usability of treetensor in several Deep Reinforcement Learning (DRL) algorithms. 
Firstly, we select three sub-domain algorithms to show how treetensor can contribute to model-based RL (MuZero), multi-agent RL (WQMIX) and Inverse RL (TREX). We select the core training functions of these three methods. Groups labeled by "(O)" is original implementation while "T" mean treetensor version.
Furthermore, we also evaluate treetensor on one of the most complicated DRL project: AlphaStar \cite{arulkumaran_alphastar_2019}, which can be thought of as a superset of several data and algorithm applications. 
It covers a variety of data types and structures, including 2D spatial observation (similar with Atari \cite{montfort_2009_atari} and Procgen \cite{procgen}), scalar and vector obs (same as mujoco \cite{mujoco}), variable-length entity tensor that shows more drastic changes in shape, discrete and continuous action space with complex relationship among different action arguments, and other structured information like several pseudo reward, etc.
Specificially, we select \textit{collate fn} in AlphaStar's training, which includes stacking, padding and pre-processing operations on above-mentioned complex data structure before a training iteration. The original code and over-written code with treetensor are shown in Appendix H. 
In all the four settings, we utilize a series of software engineering metrics \cite{mccabe_1976_cc, halstead_1977_hv, oman_1992_mi} to evaluate the code quality, including code complexity, extensibility and readability, and also report the detailed runtime, which are shown in Table \ref{table-extra-cases}.
Besides, we also show some possible applications of treetensor in Appendix I.

\begin{table*}[htb]
\renewcommand\arraystretch{1.2}
\centering
\begin{tabular}{lccccc}
\hline
\multicolumn{1}{c}{Algorithm/Metric} & \begin{tabular}[c]{@{}c@{}}Lines of Code\end{tabular} & \begin{tabular}[c]{@{}c@{}}Cyclomatic\\Complexity\end{tabular} & \begin{tabular}[c]{@{}c@{}}Halstead\\ Volume\end{tabular} & \begin{tabular}[c]{@{}c@{}}Maintainability \\
Index\end{tabular} & Runtime (ms)\\ \hline
AS collate (O)                  & 177                                                 & D (28.0)                                                        & 758                                                      & 40.8                                                           & 114.5$\pm$14.3 \\ 
\textbf{AS collate (T)}                  & \textbf{66}                                                 & \textbf{B (7.6)}                                                        & \textbf{173}                                                      & \textbf{60.7}                                     & \textbf{109.1$\pm$3.6}                       \\ \hline
MuZero \cite{schrittwieser2020mastering} (O)                  & 227                                                 & C (17.0)                                                        & 1139                                                      & 43.5                                                 & 79.1$\pm$5.4           \\ 

\textbf{MuZero (T)}         & \textbf{81}                                         & \textbf{A (4.5)}                                                & \textbf{306}                                              & \textbf{68.8}        & \textbf{71.2$\pm$6.2 }                                          \\ \hline
WQMIX \cite{rashid2020weighted} (O)                   & 287                                                 & C (11.0)                                                        & 712                                                       & 54.4                                                       & 32.3$\pm$3.9      \\ 
\textbf{WQMIX (T)}          & \textbf{123}                                        & \textbf{A (3.5)}                                                & \textbf{304}                                              & \textbf{71.3}                                                 & \textbf{30.8$\pm$3.8}  \\ \hline
TREX \cite{brown2019extrapolating} (O)                    & 309                                                 & B (8.0)                                                         & 505                                                       & 55.5                                                & \textbf{21.5$\pm$2.9}            \\ 
\textbf{TREX (T)}           & \textbf{187}                                        & \textbf{A (4.5)}                                                & \textbf{231}                                              & \textbf{77.2}                                               & 22.5$\pm$3.0    \\ \hline
\end{tabular}
 \vspace{-10pt}
\caption{Code quality metrics and runtime latency for implementing different DRL algorithms with or without treetensor. We test several algorithms in different research domains, including the most complicated case: "AS collate", which means \textit{collate function} in AlphaStar. Groups labeled by "(O)" is original implementation while "T" mean treetensor version}
\label{table-extra-cases}
 \vspace{-12pt}
\end{table*}

\subsubsection{Efficiency Benchmarks}
\label{subsection-show-benchmark}
\begin{figure}[htb]
    \centering
    \subfigure[Time Cost of Stack Operation.]{
        \centering
        \label{fig-base-operation-stack}
        \includegraphics[width=0.45\textwidth]{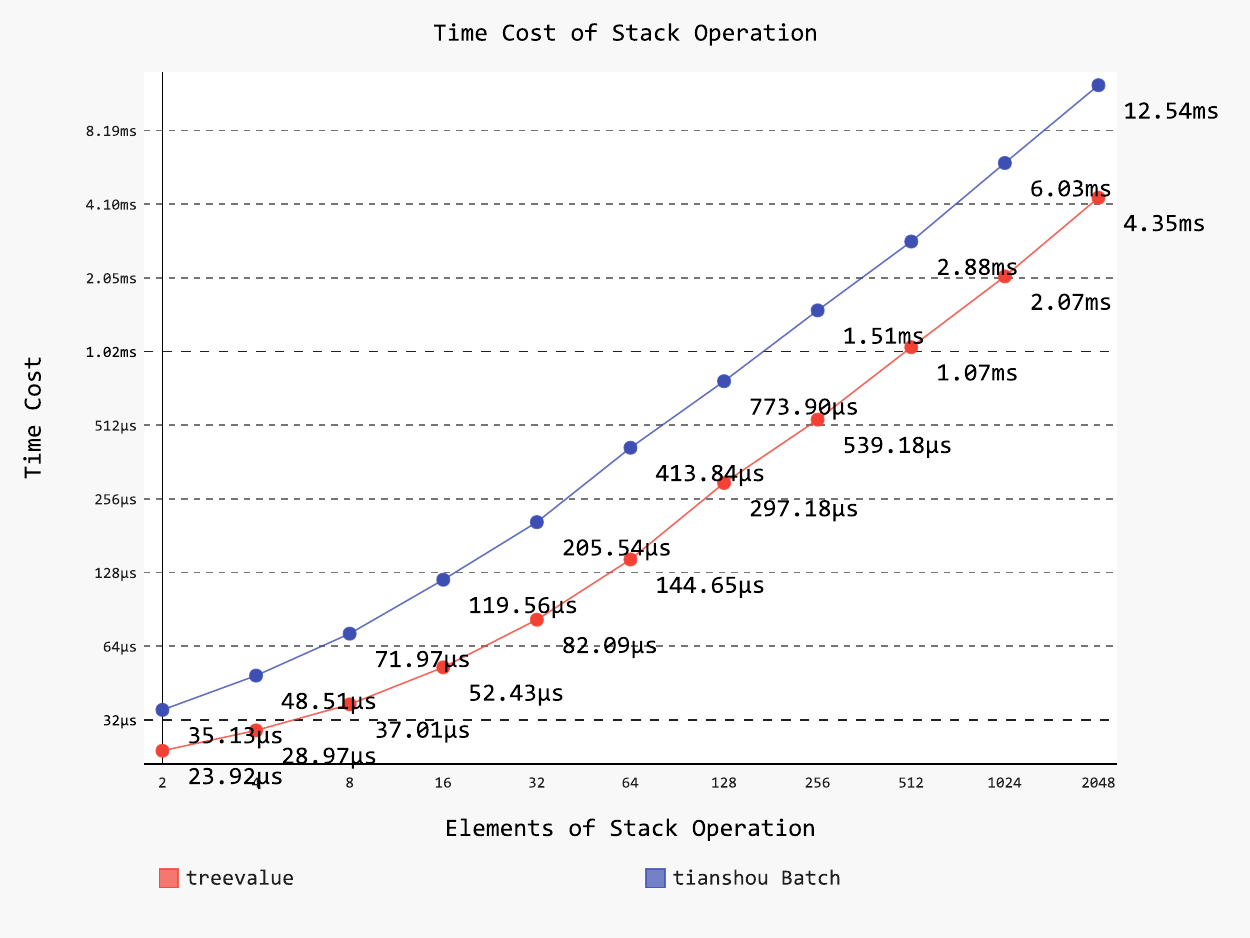}
    }
    \caption{Speed Performance Comparison of Stack and Split Operations Between treetensor and Tianshou Batch with incresing elements. The coordinates are logarithmic coordinates, which means that one grid on the y-axis means a double gap.}
    \label{fig-base-operation-stack-split}
    \vspace{-15pt}
\end{figure}

On the other hand, we also surprisingly find our implementation of treetensor shows comparable even much better than other similar libraries, like Tianshou's Batch \cite{weng_tianshou_2021}. Specifically, we test all major basic operations, such as \textit{get}, \textit{set}, \textit{init} and \textit{deepcopy} and some PyTorch tensor operations like \textit{stack}, \textit{cat} and \textit{split}. The results shown in Table \ref{tab:detailed-benchmark-with-tianshou-batch} and Figure \ref{fig-base-operation-stack-split} demonstrate that treetensor outperforms Batch with a significant margin, especially when facing larger data size and dimension. 
Also, more efficiency benchmark results can be found in Appendix B.
Furthermore, treetensor is designed for general use and offers a wide range of benefits in different aspects, runtime efficiency improvement is just a extra bonus.

\begin{table}[phtb]
\centering
\begin{tabular}{c|cc}
\hline
Operations & treetensor                  & Tianshou Batch              \\ \hline
get        & 51.6 ns ± 0.609 ns          & \textbf{43.2 ns ± 0.698 ns} \\
set        & \textbf{64.4 ns ± 0.564 ns} & 396 ns ± 8.99 ns            \\
init       & \textbf{750 ns ± 14.2 ns}   & 11.1 µs ± 277 ns            \\
deepcopy   & \textbf{88.9 µs ± 887 ns}   & 89 µs ± 1.42 µs             \\
stack      & \textbf{50.2 µs ± 771 ns}   & 119 µs ± 1.1 µs             \\
cat        & \textbf{40.3 µs ± 1.08 µs}  & 194 µs ± 1.81 µs            \\
split      & \textbf{62 µs ± 1.2 µs}     & 653 µs ± 17.8 µs            \\ \hline
\end{tabular}
\vspace{8px}
\caption{Detailed Benchmark Comparison Between treetensor and Tianshou Batch. In most operations, treetensor has better performance. Only in the get operation does Tianshou Batch outperform treetensor. This is because Tianshou Batch does not require the data structure to allow dynamic attributes, and it does not supply a large number of dynamic characteristics.}
\label{tab:detailed-benchmark-with-tianshou-batch}
   \vspace{-15pt}
\end{table}

%% file: sections/5.limitations.tex
\section{Limitations}
\label{section-limitations}


\paragraph{Weak Relation.} In fact, TreeTensor is a loosely coupled data structure, according to the definition described in Section \ref{subsection-method-definition}, the nodes do not store their parent nodes. This means that the nodes can't be a fixed component of the tree structure, and a node may be the child of different trees at the same time. Therefore, the tree nodes can only be used as a one-way index, and the value nodes can only be used as a value carrier, locality maintenance operations like that in segment trees are not able to be implemented in TreeTensor.

\paragraph{Possibility of Risky Constraint.} At the end of Section \ref{subsubsection-method-constraint-tree}, we mentioned that objects stored in a TreeTensor can be modified in-place. Among them, in the PyTorch library, this situation is very common, but the automatic validation provided by TreeTensor based on the constraint tree will not be able to capture this type of state change. Although we provide a manual full validation method for this, the abuse of this method will inevitably have a negative impact on performance. Therefore, it is necessary to pay attention when setting constraints. If the constraint can be broken by an in-place operation, it is called a "risky constraint", and such constraint should be avoided as much as possible.

%% file: sections/6.conclusion.tex
\section{Conclusion and Discusiion}
\label{section-conclusion}


In this work, we present TreeTensor, a nested tensor data structure which aims to enhance the efficiency of machine learning programming. The discoveries reveal that TreeTensor and its underlying architecture can significantly reduce the complexity of machine learning algorithm development application deployment, while preserving superior running speed to other tree data structure libraries.
Beside, the relevant toolkits of TreeTensor can also support more functions from the standpoint of simplified operation. Furthermore, for scalability, it is conceivable to increase TreeTensor's support for existing libraries, classes, and functions by improving the extension method for classes and attempting to fully realize the extension capability of modules.